\documentclass[conference]{IEEEtran}
\IEEEoverridecommandlockouts
\usepackage{cite}
\usepackage{amsmath,amssymb,amsfonts}
\usepackage{accents}
\usepackage{bm}
\usepackage[linesnumbered,ruled]{algorithm2e}
\usepackage[noend]{algpseudocode}
\usepackage{graphicx}
\usepackage{textcomp}
\usepackage{booktabs}
\usepackage{pgfplots}
\pgfplotsset{compat=1.18} 

\usepackage{multicol,multirow}
\usepackage[T1]{fontenc}
\def\BibTeX{{\rm B\kern-.05em{\sc i\kern-.025em b}\kern-.08em
    T\kern-.1667em\lower.7ex\hbox{E}\kern-.125emX}}
\usepackage{todonotes}

\newsavebox{\measurebox}
\usepackage{threeparttable}
\usepackage[font=small]{caption}
\usepackage{enumitem}
\usepackage{subfigure}
\usepackage{dblfloatfix}
\usepackage{pgfplots}
\pgfplotsset{compat=1.18} 
\usepackage{caption}

\usepackage{hyperref}
\hypersetup{
    colorlinks=false
}

\newcommand{\cheng}[1]{{{\color{purple}#1}}{}}

\newcommand*{\vsepfbox}[1]{%
  \begingroup
    \sbox0{\fbox{#1}}%
    \setlength{\fboxrule}{0pt}%
    \mbox{\kern-\fboxsep\fbox{\unhbox0}\kern-\fboxsep}%
  \endgroup
}

\begin{document}

\title{Revisiting Adversarial Perception Attacks and Defense Methods on Autonomous Driving Systems}

\author{
\IEEEauthorblockN{{\fontsize{10.5}{12.6}\selectfont 
Cheng Chen, Yuhong Wang, Nafis S Munir, Xiangwei Zhou, Xugui Zhou}}
\IEEEauthorblockA{
    Louisiana State University, Baton Rouge, LA 70803 \{cchen72, ywan248, nmunir1, xwzhou, xuguizhou\}@lsu.edu \\
    }
}

\maketitle

\begin{abstract}
Autonomous driving systems (ADS) increasingly rely on deep learning-based perception models, which remain vulnerable to adversarial attacks. In this paper, we revisit adversarial attacks and defense methods, focusing on road sign recognition and lead object detection and prediction (e.g., relative distance). Using a Level-2 production ADS, OpenPilot by Comma.ai, and the widely adopted YOLO model, we systematically examine the impact of adversarial perturbations and assess defense techniques, including adversarial training, image processing, contrastive learning, and diffusion models. Our experiments highlight both the strengths and limitations of these methods in mitigating complex attacks. Through targeted evaluations of model robustness, we aim to provide deeper insights into the vulnerabilities of ADS perception systems and contribute guidance for developing more resilient defense strategies. [Code Available at https://github.com/DepCPS/revisiting\_adversarial\_ADS]

\end{abstract}


\section{Introduction} 

Autonomous driving systems (ADS) are at the forefront of revolutionizing modern mobility, transforming our roads into safer and more efficient pathways for the future \cite{zhao2024autonomous}. Every major car manufacturer has embraced this wave by developing their own variants of autonomous driving or assisted driving solutions (e.g., Tesla's Full Self-Driving (FSD)\cite{teslafsd} ), underscoring both the competitive and innovative spirit in the industry. The key to ADS lies in accurate and powerful environment perception modules, primarily enabled through deep learning-based perception models \cite{wen2022deep}. However, these models remain highly susceptible to adversarial attacks, deliberate perturbations designed to deceive perception systems and trigger dangerous decisions or system failures \cite{cao2019adversarial, chi2024adversarial, wu2023adversarial}. 

The challenges are substantial. As adversarial techniques grow more sophisticated, existing defense strategies struggle to keep pace, often remaining confined to specially trained models and curated datasets. This disconnect between controlled testing environments and the unpredictable conditions of real-world driving underscores the urgent need for comprehensive and realistic investigations into ADS vulnerabilities and the development of robust, scalable defense mechanisms.

Reliable traffic sign recognition is vital for maintaining situational awareness and ensuring compliance with road rules. Disruptions in this process can result in improper speed regulation or navigation errors. Moreover, accurate distance estimation is essential for tasks such as collision avoidance and autonomous navigation. Adversarial perturbations that distort distance measurements can undermine the safety margins required for secure driving. 

Motivated by these challenges, this paper systematically revisits adversarial attacks and defense strategies, integrating them into the open-source commercial ADS framework. We focus on two critical perception tasks: traffic sign classification and distance regression, using YOLOv8 \cite{ultralytics_yolov8} and OpenPilot's Supercombo model as case studies to evaluate these threats comprehensively \cite{openpilot}. We experimentally examine several adversarial techniques, including Gaussian Noise \cite{gaussian_noise_wiki}, Fast Gradient Sign Method (FGSM) \cite{FGSM}, Auto Projected Gradient Descent (Auto-PGD) \cite{auto-pgd}, Simple Black-box Attack (SimBA) \cite{guo2019simpleblackboxadversarialattacks}, Robust Physical Perturbations (RP$_2$) \cite{eykholt2018robust}, and CAP-attacks \cite{zhou2024runtime}. To mitigate these vulnerabilities, we evaluate a range of defense methods, including adversarial training, image processing techniques, contrastive learning, and diffusion-based approaches. Through this comprehensive analysis, our research aims to advance the understanding of adversarial risks in ADS and support the development of more resilient and trustworthy autonomous driving systems.

Our research aims to deepen the understanding of adversarial vulnerabilities in ADS perception models and offer actionable insights into effective defense strategies, ultimately advancing the safety and resilience of autonomous driving systems. Our main contributions include:
\begin{itemize}
    \item We systematically revisit and compare the threats and limitations of different adversarial attack algorithms.
    \item We evaluate existing defense mechanisms in conjunction with different attack types to assess their effectiveness in ADS contexts.
    \item We propose directions and perspectives for advancing research on adversarial attacks and defenses in autonomous driving.
\end{itemize}

\section{Background}

\subsection{Autonomous Driving System}

Autonomous vehicles rely on sensors, actuators, advanced algorithms, often driven by machine learning, and high-performance processors to execute software. They construct and maintain a perceptual map of their surroundings using diverse sensors embedded throughout the vehicle \cite{yeong2021sensor}. For instance, radar sensors monitor the position of nearby vehicles \cite{ward2016vehicle}, while cameras detect traffic lights, read road signs, track other vehicles, and identify drivable paths \cite{trafficsigndetection, trafficlightsdetection, coifman1998real}. LiDAR sensors emit light pulses to measure distances, detect road boundaries, and recognize lane markings \cite{lidarreview}. These multimodal inputs are processed by complex models that plan driving paths and issue commands to actuators responsible for acceleration, braking, and steering.

\subsection{Adversarial Attack}

Adversarial attacks are deliberate manipulations designed to mislead machine learning models by introducing subtle, carefully crafted perturbations to input data \cite{wiyatno2019adversarial}. Although often imperceptible to humans, these perturbations can significantly degrade model performance, leading to incorrect or unsafe outputs. In safety-critical domains like autonomous driving, adversarial attacks present a substantial risk, potentially triggering hazardous behaviors or system failures \cite{badjie2024adversarial}. Identifying and mitigating these vulnerabilities is essential for ensuring the robustness and dependability of autonomous systems.

\subsection{Adversarial Defense}

Adversarial defense strategies aim to enhance the robustness of machine learning models against adversarial perturbations \cite{zantedeschi2017efficient}. Common approaches include adversarial training, where models are trained using adversarial examples; input preprocessing techniques such as image denoising and geometric transformations; and advanced representation learning methods like contrastive learning and diffusion-based generative models. Developing effective defenses is especially critical in autonomous driving to preserve the reliability and safety of perception systems under adversarial conditions.

\section{Perception Attacks}

ADS heavily relies on perception modules, which are typically built using deep neural networks to process data from various sensors. However, these perception modules are highly susceptible to adversarial attacks. In this section, we introduce several common adversarial attacks that have been extensively studied and pose significant threats to ADS.

\subsection{Gaussian Noise}

Original images, during acquisition and transmission, are often affected by noise, which degrades image quality, blurs features, and complicates analysis. Gaussian noise, characterized by a probability density that follows a Gaussian distribution, represents one of the simplest adversarial methods. This attack introduces random perturbations into the input data by adding noise sampled from a Gaussian distribution:
\begin{align}
x_{adv} = x + \epsilon, \quad \epsilon \sim \mathcal{N}(0, \sigma^2).
\end{align}
where $x_{adv}$ is the adversarial example, $x$ is the original input, and $\epsilon$ is additive noise drawn from a zero-mean normal distribution with variance $\sigma^2$. Although Gaussian noise is not specifically optimized against the model, it can notably degrade performance, particularly in environments with sensor uncertainties such as nighttime driving, fog, or rain, thereby exposing perception model vulnerabilities.

\subsection{Fast Gradient Sign Method}
The FGSM, proposed by Goodfellow et al. \cite{FGSM}, is a white-box attack that generates adversarial examples using gradient information from the neural network. By adding small, directionally aligned perturbations to the input, the model is tricked into incorrect predictions:
\begin{align}
    x_{adv} = x + \epsilon \cdot \text{sign}(\nabla_x J(\theta, x, y)).
\end{align}
where $J(\theta, x, y)$ is the model’s loss function with respect to the parameters $\theta$, input $x$, and true label $y$. The gradient sign indicates the direction in input space that maximally increases the loss. Due to its computational efficiency, FGSM is widely used for preliminary robustness assessments.

\subsection{Auto Projected Gradient Descent}
Auto-PGD \cite{auto-pgd} extends gradient-based attacks by applying iterative updates and adaptive step sizes to improve perturbations. Each iteration calculates a perturbation step and projects it back into a feasible region, maintaining imperceptibility:
{
\small
\begin{align}
    x^{t+1}_{adv} = \text{Proj}_{x + S}\left(x^{t}_{adv} + \alpha \cdot \text{sign}\left(\nabla_{x^{t}_{adv}} J(\theta, x^{t}_{adv}, y)\right)\right).
\end{align}
}
In this formulation, $S$ defines the permissible perturbation space, $\alpha$ is the step size, and $\text{Proj}$ ensures the perturbation remains within a perceptual bound. Compared with simpler methods like FGSM, Auto-PGD produces much stronger adversarial examples, making it a powerful tool for evaluating the resilience of high-performance ADS perception systems.

\subsection{Simple Black-box Attack}

SimBA, introduced by \cite{guo2019simpleblackboxadversarialattacks}, is a query-efficient black-box adversarial attack that operates without gradient estimation. It iteratively refines an adversarial perturbation $\delta$ through random sampling from an orthonormal basis (e.g., pixel or DCT frequency basis). In each step, a basis vector $\mathbf{q}$ is chosen, and $\delta$ is updated by adding or subtracting a small step $\varepsilon$ along $\mathbf{q}$. The update direction depends on output probabilities: in untargeted attacks, the direction that reduces the true class probability $p(y_{\text{true}} \mid \mathbf{x} + \delta)$ is selected; in targeted attacks, the direction that increases the target class probability $p(y_{\text{target}} \mid \mathbf{x} + \delta)$ is chosen.

The cumulative perturbation after $T$ steps is bounded by:
\begin{align}
    \|\delta_T\|_2^2 \leq T \varepsilon^2.
\end{align}
This bound ensures controlled perturbation growth. Despite its simplicity, SimBA’s orthonormal sampling strategy enables high query efficiency, making it a strong baseline among black-box attacks.

\subsection{Optimization-based Methods}
\subsubsection{Robust Physical Perturbations}

To assess deep neural network (DNN) vulnerabilities under realistic conditions, RP$_2$ \cite{eykholt2018robust} generates physical-world adversarial perturbations that remain effective across diverse environments, varying viewpoints, lighting, distances, and sensor noise. Given a classifier $f\theta(x)$ and true label $y$, the goal is to find a perturbation $\delta$ such that the perturbed input $x' = x + \delta$ is classified as a specific target label $y^* \neq y$, i.e.,
\begin{align}
    f_\theta(x + \delta) = y^*.
\end{align}

The optimization problem solved by RP$_2$ is formulated as:
\begin{align}
    \arg\min_{\delta} \ & \lambda \|\mathbf{M}_x \cdot \delta\|_p + \text{NPS} \nonumber \\
    & + \mathbb{E}_{x_i \sim \mathcal{X}_V} \left[ J\left(f_\theta(x_i + T_i(\mathbf{M}_x \cdot \delta)), y^* \right) \right].
\end{align}
Here, $|\mathbf{M}_x \cdot \delta|_p$ constrains the perturbation within the object’s surface (e.g., a stop sign) using a binary mask $\mathbf{M}_x$, and $\text{NPS}$ (Non-Printability Score) penalizes non-reproducible colors. $\mathcal{X}_V$ represents images under varying conditions, and $T_i$ denotes transformations (e.g., rotation, scaling). The function $J(\cdot)$ is typically cross-entropy loss. These perturbations highlight the real-world feasibility of adversarial threats and their implications for autonomous driving safety.

\subsubsection{CAP-Attack}

CAP-Attack \cite{zhou2024runtime} introduces a runtime adversarial patch generation technique for DNN-based Adaptive Cruise Control (ACC) systems. Unlike offline adversarial generation, this approach performs real-time pixel-level tuning on the front vehicle's image region to mislead distance prediction models: 
\begin{align}
    \min_{\Delta_t} \sum_{d \in RD_t} -\nabla g(d, \theta) + \lambda \|\Delta_t\|_p.
\end{align}

An attribution mechanism pinpoints regions most sensitive to predictions and disturbances $\Delta_t$ are confined to the bounding box of the front vehicle to reduce computation and enhance stealth. The algorithm inherits and adapts perturbations frame-by-frame, adjusting based on the displacement and size of the vehicle, ensuring both temporal coherence and concealment. This continuous optimization enhances attack effectiveness while making detection by traditional safety systems more difficult. Here, $\lambda$ controls the trade-off between the impact and invisibility of the patch, and $\|\Delta_t\|_p$ regularizes the perturbation magnitude for stealthiness.

\section{Defense Methods}

The robustness of DNNs in ADS is critical, yet adversarial attacks threaten their reliability and safety. Developing effective defense strategies to bolster adversarial robustness is therefore essential; this section outlines several widely used adversarial defense methods.

\subsection{Image Processing}

Applying simple and consistent preprocessing techniques to images before feeding them into the model is a straightforward and effective strategy. In this subsection, we introduce three such image processing techniques:

Median Blurring \cite{Xu_2018} applies median filtering to suppress adversarial noise by replacing each pixel with the median of its neighborhood, effectively preserving edges while mitigating perturbations.

Bit-depth Reduction \cite{Xu_2018} reduces the precision of pixel values, thereby decreasing the effectiveness of subtle perturbations and enhancing robustness against adversarial inputs.

Randomization \cite{xie2018mitigatingadversarialeffectsrandomization} introduces randomness, such as random resizing, padding, or noise injection, during preprocessing, which disrupts adversarial perturbations and hinders consistent model exploitation.


\subsection{Adversarial Training}
Adversarial training \cheng{\cite{zhao2024adversarialtrainingsurvey}} is a widely used and powerful defense technique that improves model robustness by solving a min-max optimization problem:
\begin{equation}
\min_{\theta}\,\mathbb{E}_{(x,y)\sim\mathcal{D}}\Bigl[\max_{\|\delta\|\le\epsilon}\mathcal{L}\bigl(f_{\theta}(x+\delta),y\bigr)\Bigr].
\end{equation}

The inner maximization, which identify the worst-case perturbation \(\delta\) within a bounded region defined by \(\epsilon\)) finds the most adversarial example, while the outer minimization updates the model parameters \(\theta\) to reduce the loss \(\mathcal{L}\) on these perturbed inputs. By incorporating adversarial examples into the training process, the model learns locally stable decision boundaries. Despite its effectiveness, especially against white-box and some black-box attacks, adversarial training generally involves greater computational overhead and may also lead to reduced performance on clean inputs.

\subsection{Diffusion Model}
Diffusion models have demonstrated impressive capabilities in generating data that resembles real-world distributions. We leverage this property to repair adversarially attacked images.We employ DiffPIR \cite{zhu2023denoising_diffpir}, a restoration framework that uses a pre-trained diffusion model as a powerful generative prior to remove adversarial or noisy perturbations. Instead of training a separate denoising model, DiffPIR harnesses the score function of the diffusion process to guide image restoration while preserving fine visual details.

\begin{align}
    \bm{x}_{t-1} = &\ \sqrt{\bar{\alpha}_{t-1}} \cdot \arg\min_{\bm{x}} \| \bm{y} - H(\bm{x}) \|^2 + \rho_t \| \bm{x} - \bm{x}_0^{(t)} \|^2 \nonumber \\
    &\ + \sqrt{1 - \bar{\alpha}_{t-1}} \cdot \left( \sqrt{1 - \zeta} \cdot \hat{\bm{\epsilon}} + \sqrt{\zeta} \cdot \bm{\epsilon}_t \right).
\end{align} 

Restoration is performed via two lightweight alternating steps: 1) Denoising applies a few steps of reverse diffusion to iteratively predict a clean image from the noisy input. 2) Projection enforces data consistency through a proximal update that aligns the output with the degraded observation. By iterating between these steps, DiffPIR achieves effective adversarial defense with strong visual fidelity and minimal architectural overhead.

\subsection{Contrastive Learning}
Contrastive learning, a self-supervised learning paradigm, can bolster robustness against adversarial attacks by learning discriminative and invariant representations. It trains models to bring augmented views of the same instance closer in the embedding space, while pushing apart unrelated instances, often via a noise contrastive estimation (NCE) loss \cite{chen2020simpleframeworkcontrastivelearning}.

Given a batch \( \mathcal{B} = \{x_i\}_{i=1}^N \), we generate two augmented views per sample \( (\tilde{x}_i, \tilde{x}_i') \) and optimize the {InfoNCE loss}:
{
\small
\begin{align}
\mathcal{L}_{\text{contrast}} = -\frac{1}{N} \sum_{i=1}^N \log \frac{\exp\left(\text{sim}(\mathbf{z}_i, \mathbf{z}_i') / \tau\right)}{\sum_{k=1}^K \mathbb{1}_{[k \neq i]} \exp\left(\text{sim}(\mathbf{z}_i, \mathbf{z}_k) / \tau\right)}
\end{align}
}
where,
\(\mathbf{z}_i = g_\phi(f_\theta(\tilde{x}_i))\) is the projected embedding, with \(f_\theta\) as the encoder (e.g., YOLOv8 backbone) and \(g_\phi\) as the projection head (MLP);
\(\text{sim}(\mathbf{u}, \mathbf{v}) = \mathbf{u}^\top \mathbf{v} / (\|\mathbf{u}\| \|\mathbf{v}\|)\) denotes cosine similarity;
\(\tau > 0\) is a temperature hyperparameter controlling the softmax sharpness;
\(K = |\mathcal{B}| - 1\) is the number of negative pairs per anchor (in-batch negatives);
\(\mathbb{1}_{[k \neq i]}\) ensures only true negatives are considered in the denominator.
We conducted quantitative evaluations of our contrastive learning-enhanced YOLOv8n model on the Traffic Signs Detection dataset.

\section{Experimental Evaluation}

\subsection{Dataset}

\begin{figure}[t]
    \centering
    \includegraphics[width=.5\linewidth]{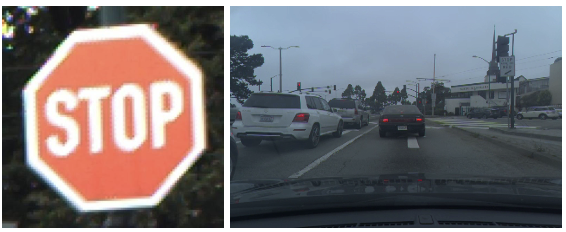}
    \caption{Example of datasets.}
    \label{fig:dataset_example}
    \vspace{-2em}
\end{figure}

We conduct experimental evaluations on both classification and regression tasks. For the classification task, we use YOLOv8, a widely used model in both academia and industry \cite{ultralytics_yolov8}, and {stop sign} images from the Traffic Signs Detection dataset \cite{darabi_cardetection}. The relative distance to the leading vehicle is the key data to ensure driving safety. Therefore, for the regression task, we use the Supercombo model, an end-to-end model used on a production ADS \cite{openpilot,schmedding2024strategic,zhou2022Strategic}, with videos from the Comma2k19 dataset \cite{comma2k19} to predict the relative distance to the leading vehicle. 
An example from each dataset is shown in Fig.~\ref{fig:dataset_example}.

\subsection{Attack Evaluation}

\subsubsection{Relative Distance Prediction}
The Supercombo model in OpenPilot is a multitask deep neural network that uses camera images to perform detection of lane lines and lead objects, as well as planning of control outputs,
enabling essential functions for autonomous driving systems. We apply various attack methods to generate adversarial patches in the region of the leading vehicle in each video frame. The model’s predicted relative distances under attack are then compared to the predictions on clean images in each frame. We evaluate the average prediction error before and after the attacks across different distance ranges.


As shown in Table~\ref{tab:attack_supercombo}, Gaussian noise exhibits the weakest attack effect, with average errors across all distance ranges remaining minimal. This indicates that the model possesses a degree of robustness to random perturbations. Auto-PGD, on the other hand, produces the largest errors in every distance range, with the highest average error reaching 34.45 meters.

Across all attack methods, the prediction error is notably higher at shorter distances, particularly within 20 meters. This suggests that the model is more vulnerable to adversarial attacks at close range, likely because perturbations occupy a larger visual area when the target vehicle is nearby, thereby enhancing the attack’s effectiveness.

\begin{table}[!h]
\centering
\vspace{-0.5em}
\caption{Avg. errors at different ranges (m) under attack}
\label{tab:attack_supercombo}
{
\begin{tabular}{@{}lllll@{}}
\toprule
\multirow{2}{*}{\textbf{Attack Method}} & \multicolumn{4}{c}{\textbf{Range (m)}} \\ \cmidrule(l){2-5} 
                  &  [0, 20]  &  [20, 40]  &  [40, 60]  &  [60, 80]   \\ \midrule
Gaussian Noise & 0.30 & 0.01 & 0.03 & 0.14 \\ 
FGSM & 18.34 & 4.25 & 3.92 & 4.65 \\ 
Auto-PGD & \textbf{34.45} & \textbf{8.43} & \textbf{8.11} & \textbf{8.49} \\ 
CAP-Attack & 29.62 & 6.73 & 6.42 & 6.83 \\
\bottomrule
\end{tabular}
}
\vspace{-0.5em}
\end{table}

\subsubsection{Stop Sign Detection}

\begin{figure}[!h]
\centering
\begin{tikzpicture}
\begin{axis}[
    width=0.48\linewidth,
    height=3.6cm,
    ylabel={Score},
    ymin=0.5, ymax=1.05,
    xtick=data,
    xticklabels={None, FGSM, Auto-PGD, RP2, Gaussian, SimBA},
    xticklabel style={rotate=30, anchor=east, font=\tiny},
    tick label style={font=\tiny},
    label style={font=\scriptsize},
    legend style={
        at={(1.05,0.5)},       
        anchor=west,
        font=\tiny,
        legend columns=1,
        column sep=1ex
    },
    legend cell align={left}
]

\addplot+[
    mark=*,
    mark options={scale=1.2, fill=blue!80!black},
    thick,
    color=blue!80!black
] coordinates {
    (0, 0.9949)
    (1, 0.7265)
    (2, 0.9509)
    (3, 0.8897)
    (4, 0.7050)
    (5, 0.9584)
};
\addlegendentry{mAP50}

\addplot+[
    mark=*,
    mark options={scale=1.2, fill=orange!80!black},
    thick,
    color=orange!80!black
] coordinates {
    (0, 0.9977)
    (1, 0.8974)
    (2, 0.9900)
    (3, 0.9867)
    (4, 0.9586)
    (5, 0.9601)
};
\addlegendentry{Precision}

\addplot+[
    mark=*,
    mark options={scale=1.2, fill=cyan!60!black},
    thick,
    color=cyan!60!black
] coordinates {
    (0, 0.9856)
    (1, 0.6104)
    (2, 0.8934)
    (3, 0.8377)
    (4, 0.6306)
    (5, 0.9498)
};
\addlegendentry{Recall}

\end{axis}
\end{tikzpicture}
\vspace{-1em}
\caption{Performance of stop sign detection with or w/o attacks.}
\label{fig:stop_sign_attack_line}
\vspace{-1em}
\end{figure}
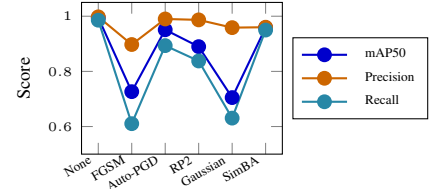

Fig.~\ref{fig:stop_sign_attack_line} presents the performance of the YOLOv8 model in detecting stop signs under different attacks. We evaluate the model using three key metrics: mAP@50, Precision, and Recall. Specifically, mAP@50 quantifies the model’s overall detection accuracy at an Intersection-over-Union (IoU) threshold of 0.5, Precision indicates the proportion of correctly identified objects among all detections, and Recall measures the model’s ability to detect all relevant stop signs.

To streamline the evaluation process, we simplify the detection task by configuring YOLOv8 for single-class detection, focusing exclusively on stop signs. As illustrated in the figure, FGSM and Gaussian Noise attacks lead to the most substantial drop in detection performance, particularly in Recall and mAP@50. Interestingly, although Auto-PGD is generally considered a stronger adversarial attack, its effectiveness appears limited in this context.

\vspace{-1em}

\subsection{Defense Evaluation}

\subsubsection{Image Processing}


To evaluate the effectiveness of classical image processing–based defenses against adversarial attacks, we conducted a series of experiments using four commonly applied input-level techniques. The results are summarized in Table~\ref{tab:image_processing_results}.

In stop sign detection, defense performance is highly dependent on the type of attack. Median Blurring is effective against simple attacks, improving mAP@50 from 70.49\% to 94.64\% under Gaussian noise and boosting precision from 89.74\% to 97.32\% under FGSM by smoothing perturbations while preserving structural edges. However, its benefits diminish under stronger attacks such as Auto-PGD, where it yields only marginal gains. In some cases, Randomization and Bit Depth Reduction even degrade performance, likely due to Auto-PGD’s already limited impact in a single-class detection task, which leaves little room for improvement and increases the risk of over-processing. Notably, Randomization underperforms under Gaussian noise, as its stochastic transformations (e.g., resizing, padding, noise injection) disrupt feature stability in already noisy inputs, reducing detection accuracy.

In regression tasks, image processing defenses also reduce adversarial prediction errors, but their effectiveness varies with object distance. Randomization is most effective at close range, reducing average error from 34.45 meters to 5.04 meters under Auto-PGD. However, beyond 40 meters, its performance drops significantly and even increases error. This is due to the sparsity and low detail of distant vehicle pixels, where random transformations can distort or erase key features, impeding depth estimation. Bit-Depth Reduction and Median Blurring offer moderate improvements, but they are less effective at close range, where stronger perturbations dominate.

These findings suggest that no single preprocessing method provides consistent robustness across different attacks and task conditions. Addressing these challenges may require combining complementary preprocessing techniques or adopting multi-model fusion strategies that account for task- and context-specific vulnerabilities.

\begin{table}[t]
\centering
\caption{{Performance} after image processing}
\label{tab:image_processing_results}
\resizebox{\columnwidth}{!}{%
\begin{tabular}{@{}lllllllll@{}}
\toprule
\textbf{Attack} & \textbf{Defense} & \multicolumn{4}{c}{\textbf{Avg. Error in Different Range (m)}} & \multicolumn{3}{c}{\textbf{Stop Sign Detection (\%)}} \\ \cmidrule(l){3-9}
\textbf{Method} & \textbf{Method} &  [0, 20]  &  [20, 40]  &  [40, 60]  &  [60, 80] & \textbf{mAP50} & \textbf{Prec.} & \textbf{Recall} \\
\midrule
\multirow{4}{*}{Gaussian} & None & 0.30 & 0.01 & 0.03 & 0.14 & 70.49 & 95.86 & 63.06 \\
                 & Median Blurring & \textbf{-0.02} & -0.12 & -0.89 & -1.89 & 94.64 & \textbf{98.47} & 89.86 \\
                 & Randomization & 4.51 & -0.21 & -11.44 & \textbf{-24.18} & 82.95 & 91.76 & 74.55 \\
                 & Bit Depth  & 1.66 & -0.53 & -2.33 & -2.57 & 70.17 & 96.88 & 62.61 \\
\midrule
\multirow{4}{*}{FGSM} & None & 18.34 & 4.25 & 3.92 & 4.65 & 72.65 & 89.74 & 61.04 \\
                      & Median Blurring & 17.60 & 6.19 & 0.66 & 1.02 & 85.49 & \textbf{97.32} & 75.00 \\
                      & Randomization & \textbf{4.62} & 0.40 & -10.23 & \textbf{-21.99} & 75.58 & 89.91 & 70.49 \\
                      & Bit Depth & 19.47 & 6.04 & -0.25 & -0.89 & 74.04 & 87.01 & 63.38 \\
\midrule
\multirow{4}{*}{Auto-PGD} & None & 34.45 & 8.43 & 8.11 & 8.49 & 95.09 & 99.00 & 89.34 \\
                      & Median Blurring & 25.57 & 6.65 & 1.37 & 2.20 & 98.84 & \textbf{99.76} & 94.80 \\
                      & Randomization & \textbf{5.04} & -0.20 & -11.56 & \textbf{-21.25} & 93.37 & 94.97 & 89.26 \\
                      & Bit Depth  & 28.87 & 7.89 & 0.75 & 0.94 & 93.82 & 98.61 & 88.06 \\
\midrule
\multirow{4}{*}{\begin{tabular}{l}
    CAP/ \\
    RP$_2$ 
\end{tabular}} & None & 29.62 & 6.73 & 6.42 & 6.83 & 88.97 & 98.67 & 83.77 \\
                & Median Blurring & 25.55 & 6.16 & 1.58 & 1.98 & 89.54 & 97.93 & 83.56\\
                & Randomization & \textbf{4.73} & 0.16 & -11.44 & \textbf{-21.38} & 87.39 & \textbf{98.93} & 81.30\\
                & Bit Depth  & 28.17 & 6.92 & 1.00 & 0.64 & 88.97 & 98.67 & 83.77 \\
\bottomrule
\end{tabular}
}
\vspace{-2em}
\end{table}

\subsubsection{Adversarial Training}
We perform adversarial training on the model using different adversarial examples as training sets, and then evaluate the retrained model under various attack scenarios. The results are summarized in Table \ref{tab:adv_training_results}.


In order to evaluate transferability, we retrained separate models using adversarial examples generated from each attack (416 stop sign images or 9600 driving video frames), and tested each retrained model on the other three adversarial examples separately.  
Additionally, to construct a diverse and representative adversarial dataset, we randomly selected 25\% of the attacked examples (104 images or 2400 frames) from each of the four attacks to be the mixed training set and randomly selected another 25\% of attacked images from each attack to be the test set.
Therefore, the mixed adversarial dataset is consistent with each individual dataset and is used to train and test another model for evaluating robustness under compound adversarial conditions. 




YOLOv8 models trained with different adversarial examples show varying levels of robustness when evaluated across multiple attack methods. Specifically, the model trained with RP2 examples performs particularly poorly against FGSM inputs, achieving the lowest scores across all evaluations, with mAP50 dropping to 40.78\%, precision to 67.01\%, and recall to 60.40\%. This sharp decline highlights a key vulnerability: RP2-trained models fail to generalize to gradient-based attacks, indicating overfitting to the RP2 attack pattern and poor cross-attack robustness. In contrast, models trained with gradient-based perturbations like FGSM or Auto-PGD exhibit more consistent and balanced performance across diverse adversarial inputs.

In relative distance prediction, we retrain the model in the same way, and adversarial training proves effective, particularly at close range. When trained with a mix of adversarial examples, the model’s average error under Auto-PGD within 0–20 meters drops sharply from 34.45 meters to just 5.84 meters. Compared to single-attack training, mixed adversarial training offers more balanced robustness across diverse attack types, highlighting its potential as a general-purpose defense strategy.
However, this improvement comes at a cost. The same model exhibits significantly larger errors at longer distances, with a maximum mean error reaching -43.04 meters. 
This behavior suggests that the adversarial loss reshapes the feature space, optimizing robustness in the near field (0–20m) but limiting generalization beyond that range. To address this, future defenses should explore distance-aware sampling, loss weighting, or multi-scale adversarial training.


\begin{table}[t]
\caption{Performance after adversarial training}
\label{tab:adv_training_results}
\resizebox{\columnwidth}{!}{%
\begin{tabular}{@{}lllllllll@{}}
\toprule
\textbf{Adversarial} & \textbf{Attack} & \multicolumn{4}{c}{\textbf{Avg. Error in Different Range (m)}} & \multicolumn{3}{c}{\textbf{Stop Sign Detection (\%)}} \\ \cmidrule(l){3-9}
\textbf{Example} & \textbf{Method} &  [0, 20]  &  [20, 40]  &  [40, 60]  &  [60, 80] & \textbf{mAP50} & \textbf{Precision} & \textbf{Recall} \\
\midrule
\multirow{4}{*}{Gaussian}
& FGSM           & 4.21 & 3.86 & 3.11 & 0.31 & 89.27 & \textbf{99.76} & 93.69 \\
& Auto-PGD       & 7.43 & 6.83 & 4.94 & 7.01 & 95.62 & 99.72 & 98.20 \\
& CAP/RP$_2$     & 5.90 & 5.69 & 3.96 & 0.80 & 83.98 & 94.70 & 85.59 \\
& Mixed & - & - & - & - & 83.32 & 97.69 & 93.85 \\
\midrule
\multirow{4}{*}{FGSM} 
& Gaussian & -0.06 & -0.08 & -0.05 & -0.20 & 94.46 & 99.32 & 98.60 \\
& Auto-PGD       & \textbf{16.83} & 4.38  & 2.77  & 3.97 & 95.13 & \textbf{99.54} & 98.39 \\
& CAP/RP$_2$     & 12.51 & 3.50  & 2.24  & 2.08 & 87.83 & 97.11 & 87.61 \\
& Mixed & - & - & - & - & 86.68 & 98.29 & 95.78 \\
\midrule
\multirow{4}{*}{Auto-PGD} 
& Gaussian & -0.09 & 0.01 & 0.10 & 0.05 & 98.85 & \textbf{99.72} & 98.65 \\
& FGSM           & 2.92  & 1.73 & 1.76 & 1.87 & 85.60 & 98.57 & 87.61 \\
& CAP/RP$_2$     & 6.50  & 2.99 & 2.34 & 2.50 & 85.61 & 97.65 & 84.40 \\
& Mixed & - & - & - & - & 82.70 & 97.88 & 91.94 \\
\midrule
\multirow{4}{*}{CAP/RP$_2$} 
& Gaussian & 0.18 & 0.53 & -0.02 & -1.43 & 55.13 & 92.20 & 54.50 \\
& FGSM           & 3.31 & 3.04 & 1.91  & -1.43 & 40.78 & 67.01 & 60.40 \\
& Auto-PGD       & 7.78 & 5.24 & 3.19  & 5.10 & 88.13 & \textbf{98.23} & 87.56 \\
& Mixed          & - & - & - & - & 67.68 & 92.44 & 74.69 \\
\midrule
\multirow{5}{*}{Mixed} 
& Gaussian & 0.25 & 0.02 & -0.02 & \textbf{-43.04} & 90.57 & 98.62 & 96.70 \\
& FGSM           & 4.17 & 1.61 & 1.41  & 0.16 & 88.63 & 98.14 & 95.19 \\
& Auto-PGD       & 5.84 & 2.91 & 3.42  & 1.47 & 87.90 & 99.09 & 97.64 \\
& CAP/RP$_2$     & 4.89 & 2.39 & 2.86  & 0.79 & 81.95 & 97.88 & 88.96 \\
& Mixed & - & - & - & - & 88.47 & \textbf{99.42} & 94.92 \\
\bottomrule
\end{tabular}
}
\vspace{-2em}
\end{table}

\subsubsection{Contrastive Learning}
{
We apply contrastive learning to enhance the feature representation of a YOLOv8 model, aiming to improve robustness in stop sign detection. This self-supervised learning approach utilizes a projection head with batch normalization and dropout, and employs a multi-positive contrastive loss with a margin to promote better feature separation. The training and test sets are the same as those for adversarial training. Performance results across different attacks are summarized in Table \ref{tab:contrastive_performance}.

While contrastive learning yields some improvements, the gains are modest. This is likely due to contrastive learning’s emphasis on feature invariance, which reduces sensitivity to input variations but does not explicitly target adversarial robustness \cite{foster2021improvingtransformationinvariancecontrastive}. In contrast, adversarial training with a single attack type often leads to overfitting, improving robustness against that attack while reducing generalization to others.
Additionally, the contrastive-trained model maintains strong and consistent performance under FGSM, suggesting that feature-level invariance may offer some resilience to simpler, gradient-based perturbations.
}

\begin{table}[t]
\centering
\caption{{Performance after} contrastive learning}
\label{tab:contrastive_performance}
\resizebox{.9\columnwidth}{!}
{%
\begin{tabular}{@{}lllll@{}}
\toprule
\textbf{Adv. Example} & \textbf{Attack Method} & \textbf{mAP50 (\%)} & \textbf{Precision (\%)} & \textbf{Recall (\%)} \\
\midrule
\multirow{5}{*}{Gaussian Noise} & Clean & 99.49 & \textbf{100.00} & 97.53 \\
                & FGSM & 78.11 & 68.93 & 59.16 \\
                & Auto-PGD & 96.24 & 96.01 & 95.17 \\
                & RP$_2$ & 92.22 & 98.92 & 79.31 \\
                & SimBA & 99.41 & 99.53 & 99.05 \\
\midrule
\multirow{5}{*}{FGSM} & Clean & 99.49 & 99.06  & 98.02 \\
                & Gaussian Noise & 80.36 & 91.53 & 66.26 \\
                & Auto-PGD & 98.50 & 98.14 & 97.15 \\
                & RP$_2$ & 93.29 & 98.26 & \textbf{100.00} \\
                & SimBA & 99.41 & 99.84 & 98.37 \\

\midrule
\multirow{5}{*}{Auto-PGD}& Clean & 99.49 & 99.06 & 98.17 \\
                & Gaussian Noise & 79.82 & 98.44 & 60.47 \\
                & FGSM & 79.67 & 95.12 & 56.60 \\
                & RP$_2$ & 90.22 & 99.86 & 81.32 \\
                & SimBA & 99.10 & \textbf{99.90} & 98.51 \\
\midrule
\multirow{5}{*}{RP$_2$}& Clean & 99.39 & 99.76 & 97.01 \\
                & Gaussian Noise & 79.01 & 97.69 & 59.34 \\
                & FGSM & 75.74 & 95.79 & 53.77 \\
                & Auto-PGD & 91.02 & 93.49 & 84.01 \\
                & SimBA & 92.11 & \textbf{100.00} & 83.51 \\
\bottomrule
\multirow{5}{*}{SimBA}& Clean & 99.39 & \textbf{99.61} & 98.11 \\
                & Gaussian Noise & 79.58 & 75.34 & 53.83 \\
                & FGSM & 75.98 & 94.30 & 52.11 \\
                & Auto-PGD & 92.50 & 82.41 & 81.67 \\
                & RP$_2$ & 92.41 & 98.82 & 84.90 \\
\midrule
\end{tabular}
}
\vspace{-2em}
\end{table}

\subsubsection{Diffusion Model}
Table~\ref{tab:diffusion_model_results} presents the performance of the diffusion model in mitigating adversarial perturbations. The results demonstrate that diffusion-based image reconstruction serves as an effective defense strategy, particularly in relative distance prediction tasks. For instance, under Auto-PGD, the average prediction error within a 20-meter range drops dramatically from 34.45 meters to just 4.98 meters. However, in cases like Gaussian noise, where the original attack is relatively weak, the diffusion process can inadvertently degrade performance, introducing unnecessary changes that lead to increased prediction errors.
Similarly, at longer distances, diffusion-repaired images often result in negatively biased predictions, with average errors becoming negative. This suggests that the model systematically underestimates distances after reconstruction, possibly due to subtle distortions or overcorrections introduced by the generative process.

In stop sign detection, the YOLOv8 model achieves a precision of over 99\% across all attacks, and reaches 100\% precision under FGSM, showing the strength of diffusion models in restoring adversarially perturbed inputs for classification.



\begin{table}[th]
\caption{Performance after diffusion model cleaning}
\label{tab:diffusion_model_results}
\resizebox{\columnwidth}{!}{%
\begin{tabular}{@{}llllllll@{}}
\toprule
 \textbf{Attack} & \multicolumn{4}{c}{\textbf{Avg. Error in Different Range (m)}} & \multicolumn{3}{c}{\textbf{Stop Sign Detection (\%)}} \\ \cmidrule(l){2-8}
 \textbf{Method} &  [0, 20]  &  [20, 40]  &  [40, 60]  &  [60, 80] & \textbf{mAP50} & \textbf{Precision} & \textbf{Recall} \\
\midrule
Gaussian  & -1.06 & -1.33 & -4.32 & -5.27 & 99.45 & 99.50 & 97.99 \\
FGSM & 6.60 & -1.06 & -3.31 & -5.03 & 97.81 & \textbf{100.00} & 93.98 \\
Auto-PGD & 4.98 & -1.05 & -3.90 & -3.77 & 99.50 & 99.96 & 99.26 \\
CAP/RP$_2$ & \textbf{8.42} & -0.67 & -2.65 & -4.08 & 93.74 & 99.33 & 89.95 \\  
SimBA & - & - & - & - & 99.50 & 99.97 & 99.26 \\

\bottomrule
\end{tabular}
}
\vspace{-1em}
\end{table}

\section{Discussion}

The evaluation of defense methods indicates that no single defense approach can fully protect against all types of attacks. For instance, the Median Blurring enhances the model's robustness to a certain extent against multiple attacks, such as mitigating errors caused by FGSM and Auto-PGD attacks by more than 10 meters, but is less effective against more subtle or dynamic threats, mitigating only about 4 meters under CAP-Attack. Conversely, methods such as diffusion models and contrastive learning show promising defense potential, yet they still require further optimization to minimize their negative impact on model accuracy in the absence of attacks. 

Additionally, time overhead is another factor that must be considered. In adversarial training and contrastive learning, the model has been retrained, and they do not require additional time for processing. While in image processing methods, on average, it takes about 20ms to process each frame or image. Most notably, DiffPIR takes 1-2 seconds to repair an attacked image, which is an unacceptable overhead. Because DiffPIR wasn’t designed with the real-time scenario and is bloated, optimizing it for real-time applications deserves further study.

Not only from the experimental results, but also from the time cost, we can observe that adversarial training is a straightforward and effective defense method.
In particular, training with mixed adversarial samples achieves a better balance across different attacks but may lead to over-defense, which reduces accuracy in long-distance prediction tasks. Future research should focus on refining adversarial training strategies, particularly in terms of selecting and designing adversarial samples to strike an optimal balance between generalization performance and defensive effectiveness.


\section{Related Work}

Many studies have attempted to attack various perception components of the autonomous driving system, such as LiDAR \cite{chen2019poba, yang2021lidarattack}, traffic sign recognition \cite{wei2022trafficsign, wei2022adversarialsticker, hu2022adversarialcolorfilm, yang2020targeted}, road lane detection \cite{jing2021toogood, boloor2020attacking}, trajectory prediction \cite{muller2022physicalhijacking, cao2022advdo}, or vehicle detection \cite{tu2020physicallyrealizable, yang2020beyonddigital, wang2021dual}.
To evaluate and improve model robustness \cite{zhou2022robustness}, numerous benchmarks have been proposed \cite{cinà2024attackbenchevaluatinggradientbasedattacks, croce2020reliable} as well as many defense methods, such as high-level representation guided denoiser \cite{Liao_2018_CVPR}, convolutional sparse coding \cite{Sun_2019_CVPR}, perturbation rectifying network \cite{Akhtar_2018_CVPR}, and runtime safety monitoring \cite{zhou2023hybrid,zhou2021data} and interventions \cite{chen2025safety}. 
However, most of these strategies~\cite{ibrahum2024survey} either do not directly study the perception module or are still tailored to image classification and evaluated with an offline image dataset. In this work, we revisit the adversarial robustness of deep learning models in the context of autonomous driving for both object detection and regression tasks from attack and defense perspectives.




\section{Conclusion}
In this study, we systematically investigate the vulnerability of ADS perception models to various adversarial attacks and evaluate multiple defense strategies. Our experimental results demonstrate that adversarial attacks significantly affect the model's classification task and regression tasks, particularly at close range. Among the defense methods evaluated, different approaches show effectiveness under specific conditions,
underscoring the urgent need to enhance the robustness of ADS perception models through innovative and adaptive defense strategies.

\bibliographystyle{IEEEtran}
        
\bibliography{reference}

\end{document}